\documentclass[conference]{IEEEtran}
\IEEEoverridecommandlockouts

\usepackage[T1]{fontenc}  
\usepackage{cite}
\usepackage{amsmath,amssymb,amsfonts}
\usepackage{algorithmic}
\usepackage{graphicx}
\usepackage{textcomp}
\usepackage{booktabs} 
\usepackage{xcolor}
\def\BibTeX{{\rm B\kern-.05em{\sc i\kern-.025em b}\kern-.08em
    T\kern-.1667em\lower.7ex\hbox{E}\kern-.125emX}}

\title{Bayesian Regression for Predicting Subscription to Bank Term Deposits in Direct Marketing Campaigns}

\author{
    \IEEEauthorblockN{Muhammad Farhan Tanvir\IEEEauthorrefmark{1}, Md Maruf Hossain\IEEEauthorrefmark{2}, Md Asifuzzaman Jishan\IEEEauthorrefmark{1}}
    \IEEEauthorblockA{\IEEEauthorrefmark{1}Institute of Computer Science and Computational Science, Universität Potsdam, Germany}
    \IEEEauthorblockA{\IEEEauthorrefmark{2}Faculty of Statistics, Technische Universität Dortmund, Germany}
    Email: farhan.tanvir@uni-potsdam.de\IEEEauthorrefmark{1}, mdmaruf.hossain@tu-dortmund.de\IEEEauthorrefmark{2}, md.jishan@uni-potsdam.de\IEEEauthorrefmark{1}
}

\begin{document}

\maketitle

\begin{abstract}
In the highly competitive environment of the banking industry, it is essential to precisely forecast the behavior of customers in order to maximize the effectiveness of marketing initiatives and improve financial consequences. The purpose of this research is to examine the efficacy of logit and probit models in predicting term deposit subscriptions using a Portuguese bank\textquotesingle{s} direct marketing data. There are several demographic, economic, and behavioral characteristics in the dataset that affect the probability of subscribing. To increase model performance and provide an unbiased evaluation, the target variable was balanced, considering the inherent imbalance in the dataset. The two model\textquotesingle{s} prediction abilities were evaluated using Bayesian techniques and Leave-One-Out Cross-Validation (LOO-CV). The logit model performed better than the probit model in handling this classification problem. The results highlight the relevance of model selection when dealing with complicated decision-making processes in the financial services industry and imbalanced datasets. Findings from this study shed light on how banks can optimize their decision-making processes, improve their client segmentation, and boost their marketing campaigns by utilizing machine learning models. 
\end{abstract}


\section{Introduction}
The banking sector, particularly in the field of marketing and customer engagement, faces the ongoing challenge of predicting customer behavior, such as term deposit subscriptions. In addition, accurate forecasting of such behavior is very crucial for improving marketing efficiency, enhancing customer segmentation, and optimizing decision-making. This research investigates whether Bayesian Data Analysis can enhance the predictive precision of term deposit subscriptions in direct marketing campaigns.

Previous studies have shown various machine learning and statistical techniques implemented to predict customer behavior in the banking industry \cite{Jiang2022}. While these approaches have shown promise, they cannot often account for data imbalances and cannot incorporate prior knowledge into the modeling process \cite{Mundargi2023}. For instance, traditional logistic regression treats data as stationary and has limited efficacy in reflecting the uncertainty inherent in financial data even if it offers reasonable predictive ability \cite{Setiyani2022}. On the other hand, machine learning models may have problems with being hard to understand and not being able to incorporate expert knowledge or adapt to real-time data dynamically \cite{Alexandra2021}. Bayesian methods are very good for financial applications \cite{Elsalamony2013} because they are flexible and can use previous information to make constantly updated predictions based on new data. Even though they have benefits, Bayesian methods for predicting customer term deposit subscriptions have not been studied much \cite{Gupta2021}. The study aims to fill that gap by using Bayesian models to make more accurate and easy-to-understand predictions. 

The comparison of two Bayesian models, Bayesian Logistic Regression and Bayesian Probit Regression, which were applied to a real-world direct marketing dataset from a Portuguese bank, is the primary contribution that this research study makes. With a total of 41,188 observations and 21 variables, the dataset was used to select a balanced subset of 10,000 records for the purpose of developing a model. Both models were able to achieve convergence successfully, and their predictive accuracy was thoroughly evaluated by employing techniques such as Leave-One-Out Cross-Validation (LOO-CV) and Out-of-Sample Prediction. 

The structure of this research paper is as follows: Section 2 presents the literature review, Section 3 covers the dataset and data quality, Section 4 describes the methodology, and Section 5 presents the results and discussion. The conclusion outlines the main conclusions and makes recommendations for further study.

\section{Literature Review}

Several studies have explored predictive techniques to enhance customer behavior forecasting in bank marketing campaigns. One study \cite{RW01} analyzed customer patterns to improve a Portuguese bank's direct marketing strategy using logistic regression and decision tree models, finding logistic regression to be more accurate. Another study \cite{RW02} applied oversampling techniques like SMOTE, ADASYN, ROS, and others to predict term deposit subscriptions, with the Naive Bayes classifier performing best, while AHC outperformed other methods. Similarly, \cite{RW03} identified "duration" as the key factor influencing term deposit decisions, with Logistic Regression and Bayesian Logistic Regression achieving 89\% and 90\% accuracy, respectively. In \cite{RW04}, XGBoost delivered the highest accuracy (91.73\%), followed by logistic regression (88.79\%) and decision trees (91.2\%).

In this study \cite{RW05}, clustering methods like K-Means, K-Medoids, and DBSCAN were combined with classification algorithms such as Decision Tree, Naïve Bayes, and Random Forest. Random Forest achieved the highest predictive accuracy, while K-Medoids outperformed K-Means and DBSCAN in clustering performance by 3.47\% and 0.3\%, respectively. Another study \cite{RW06} introduced an ensemble classification approach with Random Forest, achieving 94.02\% accuracy, and a voting classifier that improved prediction by 1.6\%, handling class imbalance to reach 95.6\% overall accuracy. In \cite{RW07}, a predictive framework using Support Vector Machines (SVM) for a Portuguese retail bank achieved 76.48\% accuracy after hyperparameter tuning, with a specificity of 82.5\% and sensitivity of 72.4\%. Another study \cite{RW08} compared Multilayer Perceptron Neural Networks (MLPNN), TAN, Logistic Regression, and C5.0 decision trees, finding C5.0 slightly outperformed the others, with "Duration" and "Age" identified as key features.

In study \cite{RW09}, techniques like SVM, Random Forest, XGBoost, Light GBM, and Gaussian Naïve Bayes were compared for predicting client behavior, with Light GBM delivering superior accuracy and faster processing. Similarly, another study \cite{RW10} analyzed models such as SVM, Gaussian Naïve Bayes, Decision Tree, Bagging, XGBoost, and Light GBM on Portuguese bank data, finding that Bagging algorithms had the highest accuracy. In contrast, \cite{RW11} introduced a hybrid deep learning model for banking, combining Self Adaptive-Sea Lion Optimization (SA-SLnO) and Deep Belief Networks (DBN) with Recurrent Neural Networks (RNN). The SA-SLnO-RB model outperformed traditional methods, improving accuracy by 0.81\%. Lastly, \cite{RW12} evaluated statistical, resampling, and distance-based methods for unbalanced term deposit datasets, showing that cluster-based resampling significantly improved positive predictions from 37\% to 80\%, with the cluster-based undersampling K-Nearest Neighbor algorithm proving most effective.

\section{Dataset}

The dataset contains detailed direct marketing campaign data carried out by a Portuguese banking institution, consisting of 41,188 samples and 21 variables \cite{b1}. The dataset is complete without any missing values, ensuring a comprehensive and reliable source for analysis. It includes a combination of categorical and integer input variables representing various elements of marketing campaigns and client demographics. The factors consist of characteristics such as age, occupation, marital status, educational attainment, housing situation, and contact information, among others. The target variable represents customer subscription behavior to a term deposit, indicating whether a client subscribed yes or not . A significant imbalance exists in the dataset, with a majority of \textquotesingle{}no\textquotesingle{} responses. To address this issue, a balanced subsample was created using oversampling techniques, where the minority class \textquotesingle{}yes\textquotesingle{} was oversampled to match the majority class. By increasing minority representation, oversampling corrects class inequality. Replicating or synthesizing minority class samples until equitable representation is achieved. Oversampling minimizes model bias toward the majority class, improving its ability to anticipate both classes. From the original 41,188 samples, 10,000 samples were randomly chosen, and the subsample was further balanced using oversampling approach which included 5,000 positive responses and 5,000 negative responses. Bayesian models are computationally intensive, especially for large datasets. Hence, a subsample of the main data was chosen for modeling. Complex models and lots of data make Bayesian inference computationally costly. It compensates with computational resources by picking a representative subsample, allowing rapid model evaluation without compromising analysis integrity. This balancing technique enables a more robust evaluation of model performance, ensuring that class imbalance does not skew the analysis or lead to biased predictions.

\section{Methodology}
In this section, the study briefly looks at the setup of the model and the criteria selected for the model.

\subsection{Logit model}

Logistic regression, also known as the logit model, is used to explain the probability of a binary outcome with a dichotomous dependent variable (0 or 1). It transforms the linear combination of predictor variables into a probability using the logistic (sigmoid) function. The formula of the logistic function can be defined as follows \cite{b2}:

$$\pi = h(\eta) = \frac{\exp(\eta)}{1 + \exp(\eta)}$$

or (equivalently) the logit link function

$$g(\pi) = \log\left(\frac{\pi}{1 - \pi}\right) = \eta = \beta_0 + \beta_1 x_1 + \ldots + \beta_k x_k$$

where $\pi$ is the probability of success, $\beta_0$ is the intercept term and $\beta_1, \beta_2, . . . , \beta_k$ are the
coefficients associated with the predictor variables $x_1, x_2, . . . , x_k$ respectively. 

\subsection{Probit model}
Probit regression is an alternative technique for estimating the probability of a binary outcome. It uses the probit function, a cumulative distribution function (CDF) of the standard normal distribution, to link predictor variables with the probability of a binary outcome. The formulation of the probit
model is as mentioned below \cite{b2}:

$$\pi = \Phi(\eta) = \Phi(x' \beta)$$

According to the conventional normal distribution, the CDF is denoted by $\Phi$. One can determine the chance that a standard normal random variable
is either less than or equal to a given value by calculating the CDF of the standard
normal distribution $\Phi$.

\subsection{Priors and Posterior Inference}
The Bayesian method uses priors based on previous research or expertise, which influence posterior distributions after data is observed. In Bayesian logit and probit models, weakly informative normal priors centered around zero were implemented to allow flexibility in predictor variable effects. Using Markov Chain Monte Carlo (MCMC), posterior distributions were estimated, which provide probabilistic support for parameter values and predictions, unlike point estimates in frequentist methods \cite{b3}.

\subsection{Priors Assumptions}

\subsubsection{Prior Selection for Logit Model}
The logit model implemented more informative priors to reflect problem-specific knowledge. The intercept had a normal prior with a mean of 3.5 and a standard deviation of 1, indicating a strong belief about the event\textquotesingle{s} baseline probability with moderate fluctuation. The regression coefficients had a normal prior with a mean of 0 and a standard deviation of 0.5, suggesting modest but meaningful predictor effects, making the model conservative for scenarios where predictors have limited influence \cite{b4}.

\subsubsection{Prior Selection for Probit Model}
The intercept had a normal prior with a mean of 0 and a standard deviation of 5, allowing for significant fluctuations in the baseline event probability. The regression coefficients had normal priors with a mean of 0 and a standard deviation of 2, suggesting a potentially greater but uncertain influence on the outcome. This approach is suitable when predictor effects are less known or expected to deviate more from zero \cite{b4}.

\section{Result and Discussion}
This section represents the research\textquotesingle{s} outcomes. This study selected 10,000 observations from the entire dataset for the final model implementation.

\subsection{Bayesian Logit Regression Model Implementation}
The results of implementing the Bayesian logit regression model, taking into account twenty-one features, are presented for consideration. 
\subsubsection{Summary of the model}
Table 1 shows the estimated coefficients and diagnostic measures for the Bayesian Logit Regression model, highlighting predictor-outcome relationships. Positive effects are observed for the "intercept" (2.34), "marital" (0.19), and "education" (0.08), while negative effects are seen for "default" (-0.44) and "contact" (-0.45). Credible intervals reflect the uncertainty of these estimates, with narrower intervals indicating greater precision.

The Rhat values of 1.00 for all variables indicate that the Bayesian model has effectively converged. This shows that the MCMC algorithm has mixed well and the posterior distributions have stabilized. Additionally, high ESS Bulk and ESS Tail values, such as for "education" and "campaign," confirm adequate sampling and reliable posterior estimates, ensuring credible and well-converged results for the model.

\begin{table*}[!htbp]
\centering
\caption{Estimated coefficients and diagnostics for the Bayesian Logit Regression model}
\begin{tabular}{lccccccc}
\hline
\textbf{Variable} & \textbf{Estimate} & \textbf{Est.Error} & \textbf{95\% CI Lower} & \textbf{95\% CI Upper} & \textbf{Rhat} & \textbf{ESS Bulk} & \textbf{ESS Tail} \\
\hline
Intercept & 2.34 & 0.81 & 0.77 & 3.94 & 1.00 & 567 & 576 \\
age & 0.02 & 0.00 & 0.01 & 0.03 & 1.00 & 1299 & 938 \\
job & 0.01 & 0.01 & -0.01 & 0.03 & 1.01 & 2273 & 584 \\
marital & 0.19 & 0.08 & 0.03 & 0.35 & 1.01 & 1164 & 620 \\
education & 0.08 & 0.02 & 0.04 & 0.12 & 1.01 & 1708 & 818 \\
default & -0.44 & 0.13 & -0.68 & -0.19 & 1.00 & 1510 & 756 \\
housing & 0.05 & 0.04 & -0.03 & 0.14 & 1.00 & 1346 & 798 \\
loan & -0.02 & 0.06 & -0.14 & 0.09 & 1.01 & 1891 & 838 \\
contact & -0.45 & 0.14 & -0.71 & -0.16 & 1.01 & 424 & 613 \\
month & -0.12 & 0.02 & -0.16 & -0.08 & 1.00 & 1344 & 637 \\
day\_of\_week & 0.04 & 0.03 & -0.02 & 0.11 & 1.00 & 1381 & 815 \\
duration & 0.01 & 0.00 & 0.01 & 0.01 & 1.01 & 1159 & 893 \\
campaign & -0.05 & 0.03 & -0.10 & -0.00 & 1.00 & 1407 & 644 \\
pdays & -0.08 & 0.02 & -0.12 & -0.04 & 1.00 & 526 & 499 \\
previous & -0.25 & 0.13 & -0.48 & 0.00 & 1.00 & 577 & 662 \\
poutcome & 0.25 & 0.18 & -0.08 & 0.61 & 1.00 & 590 & 535 \\
emp.var.rate & -0.14 & 0.07 & -0.28 & -0.00 & 1.01 & 355 & 469 \\
cons.price.idx & -0.01 & 0.02 & -0.06 & 0.04 & 1.01 & 334 & 424 \\
cons.conf.idx & 0.01 & 0.01 & -0.01 & 0.03 & 1.01 & 517 & 691 \\
euribor3m & -0.00 & 0.00 & -0.00 & 0.00 & 1.00 & 2366 & 823 \\
nr.employed & -0.33 & 0.06 & -0.45 & -0.21 & 1.00 & 472 & 631 \\
\hline
\end{tabular}%
\end{table*}

\subsection{Bayesian Probit Regression Model Implementation}
The results of implementing the Bayesian probit regression model, taking into account twenty-one features, are presented for consideration. 
\subsubsection{Summary of the model}
Table 2 presents the estimated coefficients and diagnostics for the Bayesian Probit Regression model, showing the effects of predictor variables on the outcome. The "intercept" has a strong negative estimate (-3.62) with a narrow credible interval (-3.79 to -3.46). Small negative effects are seen for "default" (-0.04), "housing" (-0.00), and "contact" (-0.05), while "education" (0.01) and "marital" (0.01) show small positive impacts. The estimates, along with standard errors and credible intervals, provide insights into the direction, magnitude, and uncertainty of these effects. The Rhat values of 1.00 for all variables confirm successful convergence, indicating that the MCMC simulations have mixed well and the posterior distributions are stable, ensuring reliable parameter estimates.

The table reports high ESS Bulk values, such as for "age" (3528), "job" (4797), and "education" (5327), indicating precise central tendency estimates from the posterior distribution. ESS Tail values, robust for variables like "housing" (2799) and "euribor3m" (2832), confirm well-sampled extreme values. These large ESS values indicate minimal autocorrelation in the MCMC chains, ensuring the reliability of the posterior estimates.

The diagnostic metrics in Table 2 confirm the robustness and stability of the Bayesian Probit Regression model. Credible intervals reflect the uncertainty of each estimate. In contrast, the coefficient estimates, Rhat, and ESS values together highlight the reliability of the Bayesian estimation process and the effects of the predictor variables.

\begin{table*}[!htbp]
\centering
\caption{Estimated coefficients and diagnostics for the Bayesian Probit Regression model}
\begin{tabular}{lccccccc}
\hline
\textbf{Variable} & \textbf{Estimate} & \textbf{Est.Error} & \textbf{95\% CI Lower} & \textbf{95\% CI Upper} & \textbf{Rhat} & \textbf{ESS Bulk} & \textbf{ESS Tail} \\
\hline
Intercept & -3.62 & 0.08 & -3.79 & -3.46 & 1.00 & 1594 & 2029 \\
age & 0.00 & 0.00 & 0.00 & 0.00 & 1.00 & 3528 & 2531 \\
job & 0.00 & 0.00 & -0.00 & 0.00 & 1.00 & 4797 & 2675 \\
marital & 0.01 & 0.01 & -0.01 & 0.03 & 1.00 & 3316 & 2786 \\
education & 0.01 & 0.00 & 0.00 & 0.01 & 1.00 & 5327 & 2282 \\
default & -0.04 & 0.02 & -0.07 & -0.00 & 1.00 & 2038 & 2161 \\
housing & 0.00 & 0.01 & -0.01 & 0.01 & 1.00 & 4360 & 2799 \\
loan & -0.00 & 0.01 & -0.02 & 0.01 & 1.00 & 4425 & 2576 \\
contact & -0.05 & 0.02 & -0.09 & -0.02 & 1.00 & 1548 & 2059 \\
month & -0.01 & 0.00 & -0.01 & -0.00 & 1.00 & 4835 & 2970 \\
day\_of\_week & 0.00 & 0.00 & -0.00 & 0.01 & 1.00 & 4417 & 2338 \\
duration & 0.00 & 0.00 & 0.00 & 0.00 & 1.00 & 3936 & 3016 \\
campaign & -0.01 & 0.00 & -0.01 & 0.00 & 1.00 & 4624 & 2308 \\
pdays & -0.00 & 0.00 & -0.01 & 0.00 & 1.00 & 1445 & 2041 \\
previous & -0.01 & 0.01 & -0.03 & 0.02 & 1.00 & 1716 & 2506 \\
poutcome & 0.02 & 0.02 & -0.01 & 0.06 & 1.00 & 1393 & 1836 \\
emp.var.rate & -0.01 & 0.01 & -0.03 & 0.01 & 1.00 & 2081 & 2135 \\
cons.price.idx & -0.00 & 0.00 & -0.01 & 0.00 & 1.00 & 1896 & 2111 \\
cons.conf.idx & 0.00 & 0.00 & -0.00 & 0.00 & 1.00 & 2250 & 2929 \\
euribor3m & -0.00 & 0.00 & -0.00 & 0.00 & 1.00 & 4447 & 2832 \\
nr.employed & -0.03 & 0.01 & -0.04 & -0.01 & 1.00 & 2214 & 2647 \\
\hline
\end{tabular}%
\end{table*}

\subsection{Model Comparison}
Table 3 summarizes the results of a Leave-One-Out Cross-Validation (LOO-CV) comparison between the Bayesian Logit Regression model \texttt{logit\_model} and the Bayesian Probit Regression model \texttt{probit\_model}. The table presents the Expected Log Pointwise Predictive Density Difference \texttt{elpd\_diff} and its associated standard error of the difference \texttt{se\_diff}, allowing for an evaluation of these models predictive performance. In this comparison, the \texttt{logit\_model} serves as the reference model with an \texttt{elpd\_diff} of 0.0 and no associated uncertainty (\texttt{se\_diff} = 0.0). The \texttt{probit\_model}, however, exhibits a significantly negative \texttt{elpd\_diff} value of -1860.9, indicating that its predictive performance is substantially worse than the \texttt{logit\_model}. The \texttt{se\_diff} for the \texttt{probit\_model} is 49.7, indicating a degree of uncertainty in this performance difference. Despite this uncertainty, the large negative \texttt{elpd\_diff} suggests that the \texttt{logit\_model} outperforms the \texttt{probit\_model} in terms of predictive accuracy based on LOO-CV results.

\begin{table}[t]
    \centering
    \caption{LOO-CV Model Comparison Results.}
    \begin{tabular}{lcc}
        \toprule
        \hline
        \textbf{Model} & \textbf{elpd\_diff} & \textbf{se\_diff} \\
        \hline
        \midrule
        logit\_model & 0.0 & 0.0 \\
        probit\_model & -1860.9 & 49.7 \\
        \hline
        \bottomrule
    \end{tabular}
    \label{tab:loo_cv_comparison}
\end{table}

The comparison highlights that the \texttt{logit\_model} is better suited for predicting the outcome in this context, as evidenced by its higher expected log pointwise predictive density. Although the \texttt{probit\_model} shows some degree of uncertainty, the results strongly favor the \texttt{logit\_model}, indicating its superior predictive capacity in this analysis.

\subsection{Out of Sample Prediction}
Table 4 presents prediction results for the Bayesian Logit Model, summarizing key statistics for three instances. The first instance has an estimated prediction of 0.257 with a standard error of 0.437 and a 95\% CI ranging from 0 to 1, indicating substantial uncertainty. Similarly, the second instance has an estimate of 0.159 (SE 0.365) and the third 0.111 (SE 0.314), both with wide CI from 0 to 1, reflecting variability and uncertainty in the predictions.

\begin{table}[t]
\centering
\caption{ Prediction Results For Logit Model}
\begin{tabular}{ccccc}
\hline
 & \textbf{Estimate} & \textbf{Est. Error} & \textbf{Q2.5} & \textbf{Q97.5} \\ 
\hline
1 & 0.257 & 0.437 & 0 & 1 \\ 
2 & 0.159 & 0.365 & 0 & 1 \\ 
3 & 0.111 & 0.314 & 0 & 1 \\ 
\hline
\end{tabular}
\label{tab:model_predictions}
\end{table}

Table 5 provides the prediction results for the Bayesian Probit Model, detailing the estimated values (Estimate), their associated standard errors (Est. Error), and the 95\% CI (Q2.5 and Q97.5) for three predicted instances. The first instance shows an estimate of 0.638 with a standard error of 0.725, and a 95\% CI ranging from 0.5 to 0.977. This suggests a relatively high predicted value with moderate uncertainty, as reflected in the large standard error and the CI, which covers a wide range. Similarly, the second instance has an estimate of 0.636, also with a standard error of 0.720 and a CI from 0.5 to 0.977. The third instance presents a slightly lower estimate of 0.542 with a standard error of 0.629, and a narrower CI (0.5 to 0.841), indicating a bit more precision in this case compared to the others.

\begin{table}[t]
\centering
\caption{ Prediction Results For Probit Model}
\begin{tabular}{ccccc}
\hline
 & \textbf{Estimate} & \textbf{Est. Error} & \textbf{Q2.5} & \textbf{Q97.5} \\ 
\hline
1 & 0.638 & 0.725 & 0.5 & 0.977 \\ 
2 & 0.636 & 0.720 & 0.5 & 0.977 \\ 
3 & 0.542 & 0.629 & 0.5 & 0.841 \\ 
\hline
\end{tabular}
\label{tab:model_predictions}
\end{table}

\subsection{Discussion}
The research findings demonstrate that both the Bayesian Logit and Probit regression models applied to 10,000 observations provide reliable estimates for predicting outcomes, though their predictive performance varies. In both models, predictors like "marital status", "education", and "contact" show significant effects. As shown in Table 1, the Bayesian Logit Regression model converges effectively with Rhat values of 1.00 and sufficiently large ESS values, ensuring reliable parameter estimation. The Probit model on Table 2 also converged with similar stability. Moreover, Table 3 compares the two models using Leave-One-Out Cross-Validation (LOO-CV). The Logit model outperforms the Probit model, with an \texttt{elpd\_diff} of 0.0 and -1860.9 for the Probit model, indicating better predictive accuracy for the Logit model despite some uncertainty \texttt{se\_diff}. On the other hand, Out-of-sample predictions in Tables 4 and 5 reflect similar patterns of uncertainty in both models, though the Logit model consistently provides better predictive performance. Overall, the Bayesian Logit Regression model proves to be more accurate and reliable in forecasting outcomes.

\section{Conclusion}

This study used a dataset from a Portuguese bank\textquotesingle{s} direct marketing campaign to assess the performance of logit and probit models in predicting client subscriptions to bank term deposits. To ensure fair and thorough testing of the models, this research employed Bayesian modeling techniques and balanced the dataset. Using Leave-One-Out Cross-Validation (LOO-CV) as a measure of prediction accuracy, this research results showed that the logit model routinely beat the probit model. Based on these results, it seems like the logit model might work better here for classifying customer subscription behavior into two categories. In complicated fields like financial marketing, the study shows how important it is to choose the right models for imbalanced datasets. Financial institutions can improve their decision-making processes, fine-tune their marketing tactics, and better target their customers with the knowledge gathered from this research. To further enhance prediction accuracy in comparable applications, future studies could investigate other modeling techniques, such as ensemble methods or non-linear classifiers.

\vspace{12pt}

\end{document}